\def\year{2022}\relax
\documentclass[letterpaper]{article} % DO NOT CHANGE THIS
\usepackage{aaai22}  % DO NOT CHANGE THIS
\usepackage{times}  % DO NOT CHANGE THIS
\usepackage{helvet}  % DO NOT CHANGE THIS
\usepackage{courier}  % DO NOT CHANGE THIS
\usepackage[hyphens]{url}  % DO NOT CHANGE THIS
\usepackage{graphicx} % DO NOT CHANGE THIS
\usepackage{natbib}  % DO NOT CHANGE THIS AND DO NOT ADD ANY OPTIONS TO IT
\usepackage{caption} % DO NOT CHANGE THIS AND DO NOT ADD ANY OPTIONS TO IT
\DeclareCaptionStyle{ruled}{labelfont=normalfont,labelsep=colon,strut=off} % DO NOT CHANGE THIS
\usepackage{algorithm}
\usepackage{algorithmic}

\usepackage{tikz}
\usepackage{color,xcolor}
\usepackage{subfigure}
\usepackage{arydshln}
\usepackage{latexsym}
\usepackage{enumerate}
\usepackage{booktabs}
\usepackage{multirow}
\usepackage{amsmath}
\usepackage{amsfonts}
\usepackage{bm}

\usepackage{amssymb}
\usepackage{CJKutf8}

\usepackage{newfloat}
\usepackage{listings}
\floatstyle{ruled}
\newfloat{listing}{tb}{lst}{}
\floatname{listing}{Listing}

\newcommand\mrightarrow{\tikz \draw [->,>=stealth] (-0.2,-0.5) -- (0.2,-0.5);}

\newcommand\mleftarrow{\tikz \draw [->,>=stealth] (0.2,-0.5) -- (-0.2,-0.5);}
\title{Mastering the Explicit Opinion-role Interaction: Syntax-aided Neural Transition System for Unified Opinion Role Labeling}
\author {
    Shengqiong Wu,\textsuperscript{\rm 1}
    Hao Fei,\textsuperscript{\rm 1}\thanks{Corresponding author}
    Fei Li,\textsuperscript{\rm 1}
    Donghong Ji,\textsuperscript{\rm 1}
    Meishan Zhang,\textsuperscript{\rm 2}
    Yijiang Liu,\textsuperscript{\rm 1}
    Chong Teng\textsuperscript{\rm 1}
}
\affiliations {
    \textsuperscript{\rm 1} Key Laboratory of Aerospace Information Security and Trusted Computing, Ministry \\
of Education, School of Cyber Science and Engineering, Wuhan University, China\\
\textsuperscript{\rm 2} Institute of Computing and Intelligence, Harbin Institute of Technology (Shenzhen), China\\
    \{whuwsq, hao.fei, lifei\_csnlp, cslyj, tengchong, dhji\}@whu.edu.cn, mason.zms@gmail.com
}

\begin{document}

\maketitle

\begin{abstract}
Unified opinion role labeling (ORL) aims to detect all possible opinion structures of `opinion-holder-target' in one shot, given a text.
The existing transition-based unified method, unfortunately, is subject to longer opinion terms and fails to solve the term overlap issue.
Current top performance has been achieved by employing the span-based graph model, which however still suffers from both high model complexity and insufficient interaction among opinions and roles.
In this work, we investigate a novel solution by revisiting the transition architecture, and augmenting it with a pointer network (PointNet).
The framework parses out all opinion structures in linear-time complexity, meanwhile breaks through the limitation of any length of terms with PointNet.
To achieve the explicit opinion-role interactions, we further propose a unified dependency-opinion graph (UDOG), co-modeling the syntactic dependency structure and the partial opinion-role structure.
We then devise a relation-centered graph aggregator (RCGA) to encode the multi-relational UDOG, where the resulting high-order representations are used to promote the predictions in the vanilla transition system.
Our model achieves new state-of-the-art results on the MPQA benchmark.
Analyses further demonstrate the superiority of our methods on both efficacy and efficiency.\footnote{Codes at \url{https://github.com/ChocoWu/SyPtrTrans-ORL}}
\end{abstract}

\section{Introduction}

\begin{figure}[!t]
\centering
\includegraphics[width=1.0\columnwidth]{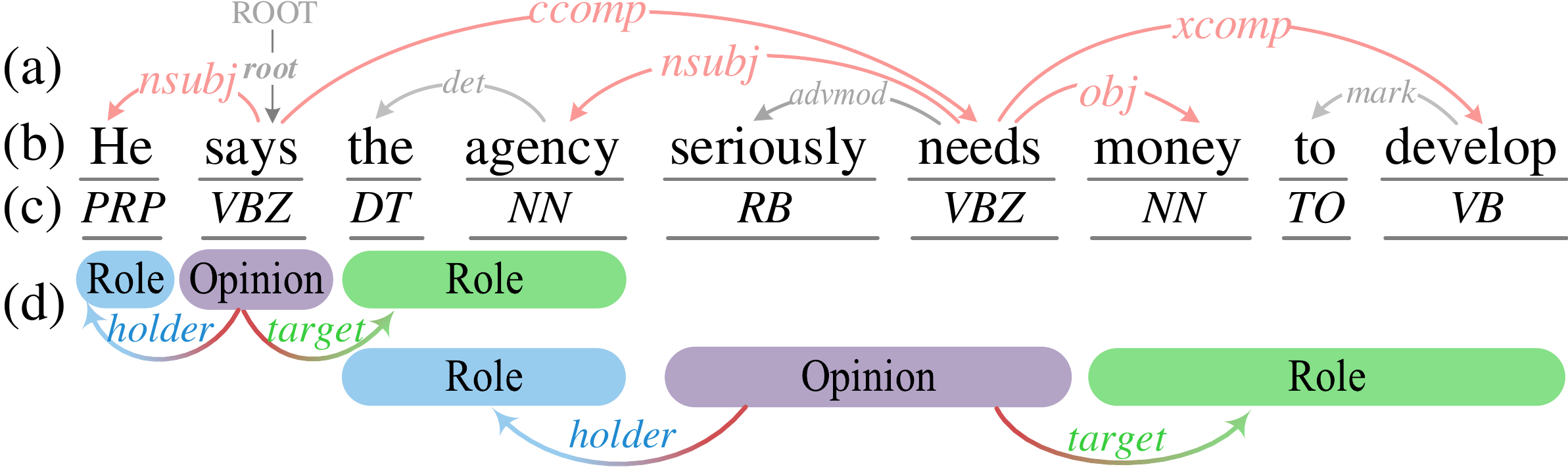}
\caption{
Illustration of the ORL structure (d) based on an example sentence (b) with its corresponding dependency structure (a) and POS tags (c).
}
\label{Example-intro}
\end{figure}

Interest in opinion role labeling (ORL) has been increasing in real-world applications as it fine-granularly analyzes users' opinions (as shown in Fig. \ref{Example-intro}), i.e., \emph{what (holder) expressed what opinions towards what (target)} \cite{kim-hovy-2006-extracting,Kannangara18,TangFYX19}.
For ORL task, traditional work adopts pipeline methods, i.e., either first extracting all terms (opinion and role terms) and then detecting the role type  (Breck et al. \citeyear{BreckCC07}; Yang et al. \citeyear{yang-cardie-2012-extracting}) or first extracting opinion terms and then finding their corresponding role terms \cite{zhang-etal-2019-enhancing,zhang-etal-2020-syntax}.
Yet, such cascade schemes can result in error propagation and meanwhile ignore the interaction between opinion and role terms.
Thereupon, follow-up studies design unified (aka. end-to-end) methods for ORL, i.e., modeling the task as an `opinion-holder-target' structure extraction via LSTM-based model (Katiyar et al. \citeyear{katiyar-cardie-2016-investigating}) or transition-based model (Zhang et al. \citeyear{ZhangWF19}).
Unified methods effectively circumvent the noise propagation and thus achieve better performances.
Nevertheless, these joint models either fail to solve the term overlap issue\footnote{
For example in Fig. \ref{Example-intro}, the role term `the agency' involves with two opinion structure simultaneously.
More generally, overlap issue refers to one word participating in multiple different span terms \cite{zeng-etal-2018-extracting,fei2020boundaries,li-etal-2021-span}.
}, one ubiquitous case in relevant data,
or stay vulnerable to the detection of lengthy mentions.

Hopefully, a recent work \cite{xia-etal-2021-unified} proposes a unified system based on the span graph model.
It works by exhaustively enumerating all possible opinion and role terms, which successfully help address the term overlap challenge (Fu et al. \citeyear{fu-etal-2021-spanner}).
Unfortunately, it still suffers from two crucial weaknesses.
The first one is the high model complexity, owing to its span enumeration nature.
Our data statistics indicate that 14.1\% of opinion and role terms span more than 8 tokens, while the terms shorter than 3 tokens account for 54.8\%.
And the span-based graph model tries to search out as many terms as possible at the cost of huge computation and time consumption.
The second issue lies in the insufficient interaction between the extraction of opinion and role terms, i.e., it takes two separate subtask decoders based on a shared encoder.
Essentially, the detection of opinion and role terms should be made inter-dependent on each other.
For example in Fig. \ref{Example-intro}, by capturing the interactions between opinions and roles, the recognition of `He' as a \emph{holder} role for the opinion `says' could meanwhile inform the detection of `the agency' as a \emph{target} role for `says', since an opinion governs both \emph{holder} and \emph{target}.
Likewise, the role as \emph{target} of `the agency' can further lead to the prediction of an opposite role (i.e., \emph{holder}) for the opinion term `seriously needs'.

In this work we try to address all the above challenges by investigating a novel unified ORL framework.
First of all, we reformalize the task as structure parsing and solve it with a neural transition model.
Different from Zhang et al. (\citeyear{ZhangWF19}), we redesign the transition architecture so as to support the recognition of overlapping terms.
The transition system produces all possible opinion structures (including the overlapped one) in one shot by retaining the advantage of linear-time complexity.
Then, we equip the transition model with a PointNet (Vinyals et al. \citeyear{VinyalsFJ15}) for determining the end boundaries of opinion terms and role terms, by which we break through the restraint of lengthy terms.

We further consider the leverage of external rich syntax knowledge for enhancing the above vanilla transition system.
On the one hand, we reinforce the boundary recognition of PointNet by integrating the linguistic part-of-speech (POS) tag features.
Intuitively, the POS pattern could help better determine the boundary of the span terms \cite{lin-etal-2018-multi-lingual}.
For instance, the phrase `the agency' with the implication of `DT-NN' POS tags is more likely to be a term.
On the other hand, we propose a dependency-guided high-order interaction mechanism to achieve explicit interactions between opinions and roles.
Essentially, the opinion-role structure echoes much with the dependency structure, as can be illustrated in Fig. \ref{Example-intro}, and such external syntax structure could promote the interactions between the opinion and role terms.
Technically, we construct a \underline{\bf u}nified \underline{\bf d}ependency-\underline{\bf o}pinion \underline{\bf g}raph (UDOG), in which we simultaneously model the dependency structure and the partially predicted opinion-role structure as one unified graph.
We especially devise a novel \underline{\bf r}elation-\underline{\bf c}entered \underline{\bf g}raph \underline{\bf a}ggregator (RCGA) to encode the multi-relational UDOG,
where the resulting global-level features will be used for promoting the discovery of the rest of unknown opinion structures.

We evaluate the efficacy of our framework on the ORL benchmark, MPQA v2.0 (Wiebe et al. \citeyear{WiebeWC05}).
Experimental results demonstrate that our method obtains new state-of-the-art results than the current best-performing baseline, meanwhile achieving faster decoding.
We further show that the leverage of the external syntax knowledge effectively promotes the term boundary recognition and the interactions between the detection of opinion-role structures.
More in-depth analyses further reveal the strengths of our framework.
We summarize the contributions of this work below:

$\bigstar$ We explore a novel end-to-end solution for ORL based on a neural transition model with a pointer network.
The design of transition with PointNet enables more accurate predictions on longer terms, and solving the overlap issue meanwhile enjoying a linear-time complexity.

$\bigstar$ We strengthen the term boundary detection with linguistic POS tags.
We also achieve the explicit interactions of opinion role terms by the use of syntactic dependency structure knowledge, by which we capture rich global-level high-order features and substantially enhance the predictions.

$\bigstar$ Our system obtains new state-of-the-art ORL results on benchmark data.
Further in-depth analyses are presented for a deep understanding of our method.

\section{Related Work}

ORL, i.e., mining the individuals' fine-grained opinion, has been a well-established task in NLP community (Hu et al. \citeyear{HuL04}; Pang et al. \citeyear{PangL07}; Chen et al. \citeyear{ChenC16}).
Earlier works mostly take two-step methods for the task \cite{BreckCC07,yang-cardie-2012-extracting,marasovic-frank-2018-srl4orl,zhang-etal-2019-enhancing}.
Recent efforts consider jointly extracting the overall opinion-role results in one shot, modeling the implicit interactions between two stages meanwhile reducing the error propagation (Katiyar et al. \citeyear{katiyar-cardie-2016-investigating}).

It is noteworthy that Zhang et al. (\citeyear{ZhangWF19}) investigate a neural transition method, achieving end-to-end task prediction while keeping a linear-time model complexity (Zhang et al. \citeyear{zhang-clark-2010-fast}; Dyer et al. \citeyear{dyer-etal-2015-transition}; Fei et al. \citeyear{0001ZLJ21}).
\textit{This work inherits the merits of the transition modeling of ORL, but ours further advances in three aspects:}
(1) we re-design the transition framework so as to address a crucial \emph{mention overlap} challenge in the task;
(2) we enhance the transition model with a pointer network for the end-boundary recognition (Fern{\'a}ndez-Gonz{\'a}lez et al. \citeyear{fernandez-2020-transition}), which significantly enables more accurate term detection, especially for those long terms.
(3) we consider making use of the opinion-role interaction for better task performances.

More recently, Xia et al. (\citeyear{xia-etal-2021-unified}) solve the overlap issue with a span-based graph model by iteratively traversing all possible text spans.
Unfortunately, such exhaustive enumerating task modeling results in huge computational costs, i.e., $\mathit{O}(n^4)$.
Meanwhile, their work fails to explicitly manage the interactions between the opinion-role structures, where further task improvements can be pretty limited.
Besides, Xia et al. (\citeyear{xia-etal-2021-unified}) implicitly integrate the syntactic constituency information based on a multi-task learning framework.
Differently, we take the advantage of dependency syntax knowledge for building the UDOG, through which we achieve the goal of explicit and sufficient opinion-role interactions, and thus obtain substantially enhanced performances.

\begin{figure*}[!t]
\centering
\includegraphics[width=0.95\textwidth]{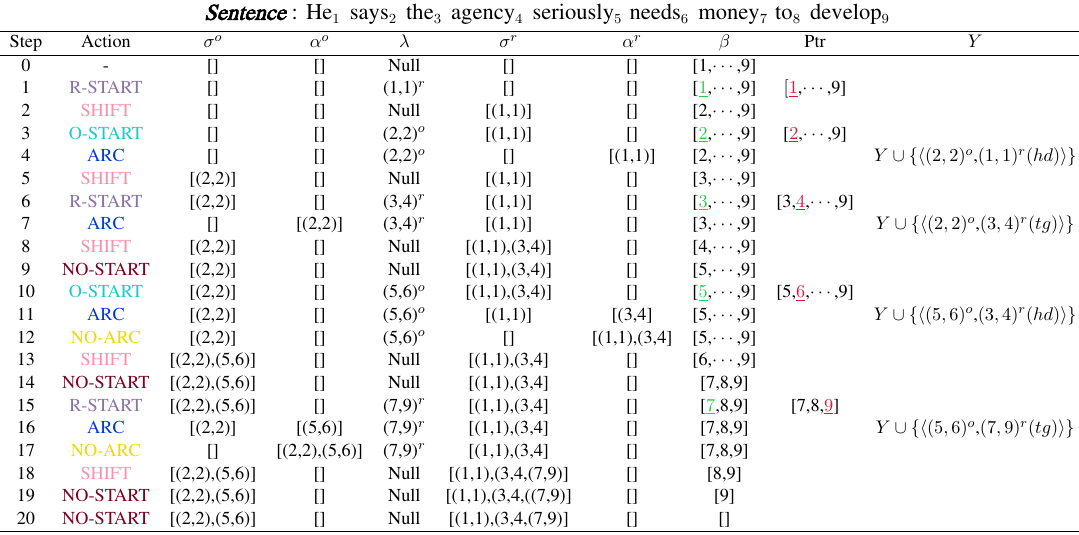}
\caption{
An illustration of the transition process. 
`Ptr' means PointNet for end boundary detection of opinion or role terms.
The underlined number in $\beta$ and `Ptr' marked with green or red denotes the start or end index of opinion or role terms, respectively.
}
\label{Transition}
\end{figure*}

\section{Transition System}

\noindent\textbf{Task Formalization.}
We model the task as an `$\langle$opinion, role(type)$\rangle$' structure parsing.
Given a sentence $X$=$\{w_1, \cdots, w_T\}$, our system outputs a list of pairs $Y$=$ \{\langle o, r(c)\rangle_q, \cdots \}_{q=1}^{Q}$,
where $o$ is an opinion term, with $o$=$\{w_i,\cdots, w_j|1$$\le$$i$$\le$$j$$\le$$T\}$;
$r$ is a role term, with $r$=$\{w_{i^{'}},\cdots, w_{j^{'}}|1$$\le$$i^{'}$$\le$$j^{'}$$\le$$T\}$;
$c$$\in$$\{hd, tg\}$ is a role type, and `$hd$' means `\emph{holder}' and `$tg$' represents `\emph{target}'.

\subsection{Transition Definition}

Overall, the transition system comprises of actions and states, where the actions determine the parsing decision and produce transition states, while the states control the action prediction and store partial outputs.

\noindent\textbf{States.}
The transition state is formally defined as  $s=(\sigma^{o},\alpha^{o},\sigma^{r},\alpha^{r}, \lambda, \beta, A, Y)$.
Specifically,  
$\sigma^{o}$ and $\sigma^{r}$ are two \emph{stacks} to store processed opinion terms and role terms, respectively.
$\alpha^{o}$ and $\alpha^{r}$ are also two \emph{stacks} for holding opinion and role terms that are popped temporarily from $\sigma^{o}$ and $\sigma^{r}$, respectively.
$\lambda$ is a \emph{variable} referring to either an opinion or role term.
$\beta$ refers to a \emph{buffer} containing unprocessed tokens in $X$.
$A$ is an action \emph{list} to record historical actions.
$Y$ stores the generated pairs.

\noindent\textbf{Actions.}
We design six actions as defined in following:
\begin{itemize}
\setlength{\topsep}{0pt}
\setlength{\itemsep}{0pt}
\setlength{\parsep}{0pt}
\setlength{\parskip}{0pt}
    \item \textbf{O-START}/\textbf{R-START} means that the top element in $\beta$ is a start token of an opinion term or a role term.
    Once determining the start position by the action, the end position will further be obtained by the PointNet,
    and then the $\lambda$ representation will be updated.

    \item \textbf{NO-START} implies that the top element in $\beta$ is not a start token of any term, so pop it out of the $\beta$.
      
    \item \textbf{ARC} builds a valid relation between the top element in $\sigma^*$ and $\lambda$, where $* \in \{o, r\}$.
    If $\lambda$ is an opinion term $o$, we assign a specific role type $c$ for the $\langle$o, r$\rangle$ pair, i.e., $\lambda$ and the top element in $\sigma^r$, and pop the top element out of $\sigma^r$ into $\alpha^r$.
    And if $\lambda$ is a role term $r$, we decide the role type $c$ for the $\langle$o, r$\rangle$ pair, i.e., the top element in $\sigma^o$ and $\lambda$, and pop the top element out of $\sigma^o$ into $\alpha^o$.

    \item \textbf{NO-ARC} means there is no valid relation between the top element in $\sigma^*(*\in\{o,r\})$ and $\lambda$.
    If $\lambda$ is an opinion term,we pop the top element out of $\sigma^r$ into $\alpha^r$; Otherwise, we pop the top element out of $\sigma^o$ into $\alpha^o$.

    \item \textbf{SHIFT} represents the end of relation detection for $\lambda$.
    So we first pop all elements out of $\alpha^*$ into $\sigma^*(* \in \{o, r\})$.
    Then, if $\lambda$ is an opinion term, we pop the element out of $\lambda$ into $\sigma^o$; otherwise we pop the element out of $\lambda$ into $\sigma^r$.
    Finally, we pop the top element out of the $\beta$.

\end{itemize}

\noindent\textbf{Action Preconditions.}
We also design several preconditions \cite{fan-etal-2020-transition} to ensure the valid action candidates to predict at each transition step $t$.
For example, NO/R/O-START can only be conducted when $\beta$ is not empty and $\lambda$ is empty.

\noindent\textbf{Linear-time Parsing.}
Since each unprocessed token in $\beta$ will be handled only once, i.e., determining whether a token is a start of the opinion or role term, our transition system takes linear-time complexity.
Fig. \ref{Transition} shows an example of how the transition system works.
At the initial state, all unprocessed words are stored in $\beta$.
Then, the states are sequentially changed by a list of actions, during which the opinion terms and their corresponding role terms are extracted into $Y$.
The procedure terminates once both buffer $\beta$ and $\lambda$ turn empty.
Finally, our system will output all pairs into $Y$, e.g., total four pairs, `$\langle$says, He($hd$)$\rangle$', `$\langle$says, the agency($tg$)$\rangle$', `$\langle$seriously needs, the agency($hd$)$\rangle$' and `$\langle$seriously needs, money to develop($tg$)$\rangle$', of the sentence in Fig. \ref{Transition}.

\section{Neural Transition Model}

We construct neural representations for transition states in $s$, based on which we employ neural networks to predict actions.
If an O/R-START action is generated, the PointNet will determine the corresponding end position.
And if an ARC action is yielded, the role type detector will assign a role type for the role term in an opinion-role pair.

\noindent\textbf{Word Representation.}
We consider two types of representation for each word $w_i$, including a word-level embedding $\bm{x}^{w}_i$ from a pre-trained model and a character-level representation $\bm{x}^{c}_i$ generated by a convolutional neural network:
\begin{equation}
\setlength\abovedisplayskip{2pt}
\setlength\belowdisplayskip{2pt}
\bm{x}_i = [\bm{x}^{w}_i;\bm{x}^{c}_i] \,.
\end{equation}
where $[;]$ denotes a concatenation operation.

\noindent\textbf{State Representation.}
We construct neural state representation $\bm{e}_t^s$ for the state $s_t$ at step $t$.
We first use a BiLSTM to encode each word in buffer $\beta$ into representation $\bm{h}_i$, and also generate the overall $\beta$ representation $\bm{e}^{\beta}_{t}$.
We use another BiLSTM to represent $\lambda$ and the action list $A$, i.e., $\bm{e}^{\lambda}_t$ and  $\bm{e}^{A}_{t}$.
We utilize two separate Stack-LSTMs to encode the stacks $\sigma^o$ and $\sigma^r$, i.e., $\bm{e}^{o}_t$ and $\bm{e}^{r}_t$.
Finally, we summarize all the above items as the overall state representation $\bm{e}_t^{s}$:
\begin{equation}
\setlength\abovedisplayskip{2pt}
\setlength\belowdisplayskip{2pt}
\bm{e}_t^{s} = [\bm{e}^{\lambda}_t; \bm{e}^{o}_t; \bm{e}^{r}_t; \bm{e}^{A}_{t}, \bm{e}^{\beta}_{t}] \,,
\end{equation}

\begin{figure}[!t]
\centering
\includegraphics[width=1.0\columnwidth]{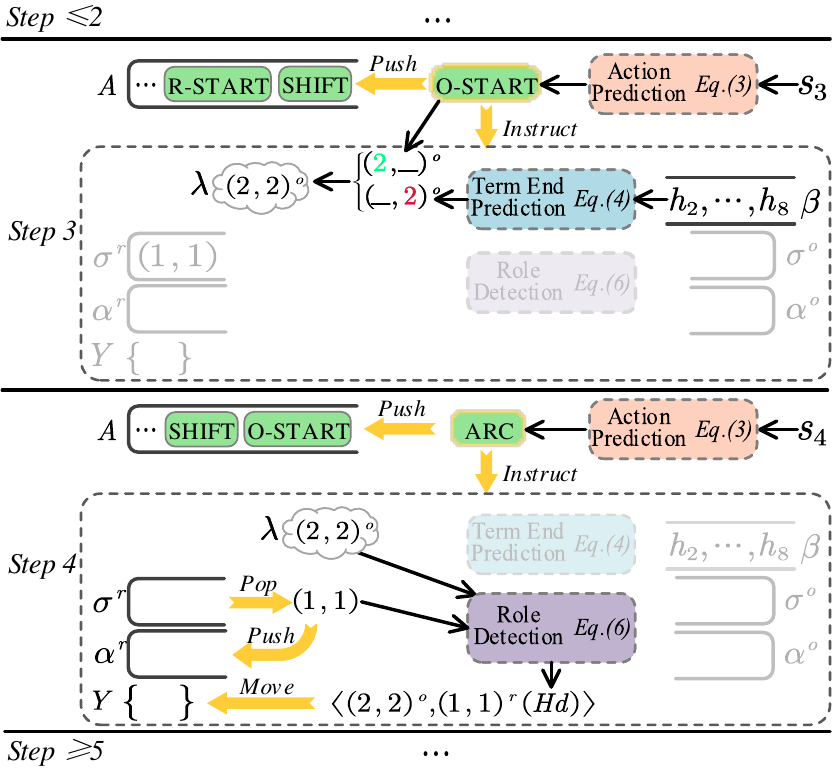}
\caption{
The illustration of transition \emph{step 3} to \emph{step 4} with the vanilla neural model.
}
\label{TPtr1}
\end{figure}

\noindent\textbf{Action Prediction.}
Based on the state representation $\bm{e}_t^{s}$, we first apply a multi-layer perceptron (MLP) on it, and then predict the action $y_t^A$ via a softmax, as shown in Fig. \ref{TPtr1}:
\begin{equation}
\setlength\abovedisplayskip{2pt}
\setlength\belowdisplayskip{2pt}
y_t^A = \mathop{\text{Softmax}}\limits_{\mathcal{A}(t)}(\text{MLP}(\bm{e}_t^{s}))\,,
\end{equation}
where $\mathcal{A}(t)$ indicates the set of valid action candidates at step $t$ according to the aforementioned preconditions.

\noindent\textbf{Term End Prediction via PointNet.}
Based on the current start index $i$ of an opinion or role term, as well as the hidden representation $\{\bm{h}_i,\cdots, \bm{h}_T\}$ from $\beta$, 
we adopt the PointNet to detect the end index $j$, as can be found in Fig. \ref{TPtr1}:
\begin{align}
\setlength\abovedisplayskip{3pt}
\setlength\belowdisplayskip{3pt}
\label{uik} u_{ik}  = \text{Tanh}(&\bm{W}_1\bm{h}_{i} + \bm{W}_2\bm{h}_k), k=[i,\cdots, T] \,, \\
\label{oik} o_{ik} &= \text{Softmax}(u_{ik}) \,, \\
\label{index j} j &= \mathop{\text{Argmax}}\limits_{i \le  j \le T}(o_{ik}) \,, 
\end{align}
where $\bm{W}_{*}$ are trainable parameters (same in below).

After obtaining the start and end index of the term, we generate its representation $\bm{a}^{*}(* \in \{o, r\})$ as follows:
\begin{equation}
\setlength\abovedisplayskip{2pt}
\setlength\belowdisplayskip{2pt}
\bm{a}^{*} = \bm{W}_3[\bm{h}_i; \bm{h}_{j}; \bm{x}^{P[i:j]}] 
\label{span-representaion}
\end{equation}
where $\bm{x}^{P[i:j]}$ is a term length embedding.
The term representation will be employed to update the $\lambda$ representation.

\noindent\textbf{Role Type Detection via Biaffine.}
At the meantime, once a valid relation is determined by an ARC action between an opinion term $o$ and a role term $r$, a biaffine function \cite{BiaffineDozatM17} assigns a type $c$ for the role (cf. Fig. \ref{TPtr1}):
\setlength\abovedisplayskip{2pt}
\setlength\belowdisplayskip{2pt}
\begin{align}
\label{biaffine} y^c =& \text{Tanh}(\left[
\begin{array}{c}
  \bm{a}^o 
\end{array}
\right]^\mathsf{T}
\cdot \bm{W}_4 \cdot \bm{a}^{r}) \,, \\
&c = \text{Softmax}(y^c) \,.
\end{align}
By now we can ensemble one complete opinion-role pair $\langle o,r(c)\rangle$, which we will add into $Y$.

\noindent\textbf{Training Object.}
The target is to produce the pair structure $\hat{Y}$ given the gold $Y$, which can be decomposed into three sub-objectives, including action prediction loss $\mathcal{L}_a$, end boundary prediction loss $\mathcal{L}_p$ and role detector loss  $\mathcal{L}_c$.
We adopt cross-entropy of these predictions over gold labels.
We can ensemble the above items:
\begin{equation}
\setlength\abovedisplayskip{3pt}
\setlength\belowdisplayskip{3pt}
\mathcal{L} = \mathcal{L}_{a} + \mathcal{L}_p + \mathcal{L}_c + \frac{ \eta}{2} \left \| \theta \right\|^2 \,,
\end{equation}
where 
$\theta$ are the overall model parameters, and $\eta$ is the coefficient of $\ell_2$-norm regularization.

\section{Syntax-enhanced Transition Model}

The above neural transition model generates a list of actions, from which the final opinion role pairs can be parsed out incrementally.
However, the vanilla system takes merely the local-level clues for each action prediction \cite{LiZP20}, meanwhile without any explicit interaction between opinion and role terms.
Therefore, in this section, we consider constructing global-level features and modeling the explicit interaction by integrating the syntax knowledge.
Based on the vanilla system, on the one hand we propose a POS-aware PointNet; on the other hand we introduce a dependency-guided high-order interaction mechanism.

\subsection{POS-aware PointNet for Term Prediction}

Since POS tags can offer potential clues for the term boundary recognition \cite{NieTSAW20,Wu0RJL21}, we therefore consider integrating such features into the PointNet.
We update the raw pointing calculation in Eq.(\ref{uik}) with
\begin{equation}\small
\setlength\abovedisplayskip{3pt}
\setlength\belowdisplayskip{3pt}
\label{pos-guided ptr} u_{ik} = \text{Tanh}(\bm{W}_5[\bm{h}_i;\bm{x}_{i}^{p}] +  \bm{W}_6 [\bm{h}_k;\bm{x}_{k}^{p}]  + \Delta\text{F}_{k-1, k, k+1}) \,,
\end{equation} 
where we first concatenate a POS embedding $\bm{x}_{i}^{p}$ with token representation $\bm{h}_i$ for $w_i$. 
$\Delta\text{F}_{k-1, k, k+1}$ is a \emph{Boundary Differential Enhancer} (BDE) that captures the boundary differences between the former and the next POS tag of token $k$:
\begin{equation}
\setlength\abovedisplayskip{3pt}
\setlength\belowdisplayskip{3pt}
\label{BDE} \Delta\text{F}_{k-1, k, k+1} =  \bm{W}_7[(\bm{x}_{k}^{p}-\bm{x}_{k-1}^{p});(\bm{x}_{k+1}^{p}-\bm{x}_{k}^{p})] \,.
\end{equation}

\subsection{Dependency-guided High-order Interaction}

Our main motivation is to encourage sufficient interactions between those already-yielded opinion-role structures by the vanilla system so as to better discover the unknown structures.
It is possible to directly model the interactions within those opinions and roles that are overlapped together.
However those latent interactions between the terms without explicit connections cannot be sufficiently modeled.
Thus, we consider leveraging the external syntactic dependency information.
Intuitively, the word-word relations depicted by the dependency tree offer much aid for building high-order interactions \cite{li-etal-2020-high}.
Practically, we first construct a unified graph structure to simultaneously model the previously yielded opinion-role structure with the dependency tree globally, i.e., unified dependency-opinion graph (UDOG).
We then propose a relation-centered graph aggregator (RCGA) to encode the UDOG, where the resulting high-order representations are further used to promote the predictions in the vanilla transition system.

\begin{figure}[!t]
\centering
\includegraphics[width=1.0\columnwidth]{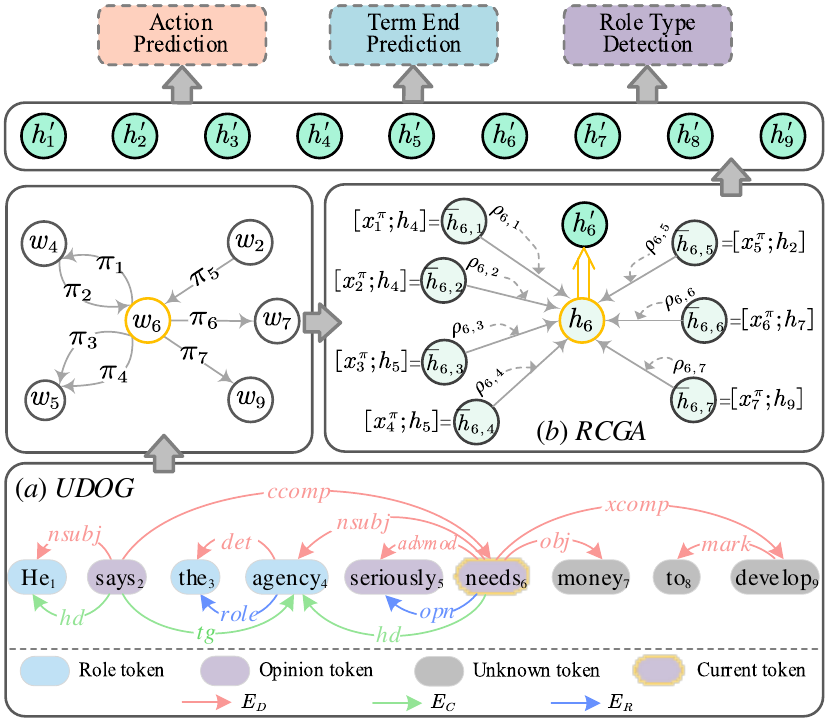}
\caption{
Encoding UDOG (a) via RCGA (b).
The resulting high-order representation is used for action prediction, term end prediction, and role type detection.
}
\label{UDOS1}
\end{figure}

\noindent\textbf{Building UDOG.}
Intuitively, the opinion-role structure coincides much with the syntactic dependency structure, and the co-modeling of these two structures would effectively help the interactions between the opinion-role pairs.
Given a sentence of words, the corresponding dependency tree, and the currently generated opinion-role structures in $Y$, the unified dependency-opinion graph can be formalized by ${G}$=$({V}, {E})$.
${V}$ is a set of word nodes $w_i$, and ${E}$ is a set of labeled edges $\pi_v$, with ${E}$=${E}_D$$\cup$${E}_C$$\cup$${E}_R$.
${E}_D$ represents the dependency edges attached with syntactic labels, e.g., `He'$^{\overset{\text{nsubj}}{\mleftarrow}}$`says'.
${E}_C$ means the inter-term links, e.g., `says'$^{\overset{\text{tg}}{\mrightarrow}}$`the agency'; while ${E}_R$ means the intra-term links\footnote{We construct the intra-term arcs by a \emph{tail-first} strategy \cite{barnes-etal-2021-structured}, connecting tail token to all the other tokens.}, e.g., `the'$^{\overset{\text{role}}{\mleftarrow}}$`agency', `seriously'$^{\overset{\text{opn}}{\mleftarrow}}$`needs', as shown in Fig. \ref{UDOS1}($a$).
Note that UDOG is a multi-relation graph, as a pair of words may be connected with more than one edge.
We actively update UDOG once a new opinion-role pair is detected out.

\noindent\textbf{Graph Encoding via RCGA.}
Existing graph encoders, e.g., graph convolutional network \cite{marcheggiani-titov-2017-encoding} or graph recurrent network \cite{zhang-zhang-2019-tree}, have been shown prominently on graph modeling, while they may fail to directly handle our UDOG due to two key characteristics: \emph{labeled edges} and \emph{multi-relational edges}.
Thus, we propose a novel relation-centered graph aggregator (RCGA).
The key idea is to take the edges as mainstays instead of the neighbor nodes.
Technically, we aggregate the information for node $w_i$ from its neighbors as well as the labeled edges:
\setlength\abovedisplayskip{2pt}
\setlength\belowdisplayskip{2pt}
\begin{align}
&\bar{\bm{h}}_{i,v} = [\bm{h}_j; \bm{x}_{i,v}] \,, \quad  j \in \mathcal{B}(i,v) \,, \\
\rho_{i,v} &= \mathop{\text{Softmax}}\limits_{v \in \mathcal{E}(i)}(\text{LeakyReLU}(\bm{W}_8[\bm{h}_i; \bar{\bm{h}}_{i,v}])) \,, \\
&\bm{h}^{'}_{i} = \frac{1}{| \mathcal{E}(i)|}\begin{matrix} \sum_{v \in \mathcal{E}(i)} \end{matrix}  \rho_{i,v} \bm{W}_{9}\bar{\bm{h}}_{i,v} \,, 
\end{align}
where $\bm{x}_{i,v}$ is the label embedding of the edge label $\pi_v$,
$v \in\mathcal{E}(i)$ is the edge directly connecting with node $w_i$, 
and $j \in\mathcal{B}(i,v)$ is the target node of edge $v$.
$\bar{\bm{h}}_{i,v}$ is the relation-centered representation that entails both neighbor node and the edge label information.
$\rho_{i,v}$ is the neighboring matrix.
In particular we take 2 iterations of RCGA propagation for a thorough structure communication.
Also, RCGA performs above calculations whenever the UDOG is updated.
Fig. \ref{UDOS1}($b$) illustrates the RCGA based on $w_6$ as the current node.

\noindent\textbf{Updating Predictions.}
Once obtaining the enhanced node representations $\{\bm{h}^{'}_{1},\cdots,\bm{h}^{'}_{T}\}$, we then update the predictions in the vanilla systems.
We first enhance the action prediction by additionally adding a global-level feature representation $\bar{\bm{g}}$ into the original $\bm{e}^s_t$:
\begin{equation}
\begin{aligned}
\setlength\abovedisplayskip{2pt}
\setlength\belowdisplayskip{2pt}
\bar{\bm{g}} = \text{Graph-Pooling}&(\{\bm{h}^{'}_1, \cdots, \bm{h}^{'}_T\}) \,, \\
\bm{e}_t^{s} := [\bm{e}_t^{s}&;\bar{\bm{g}}] \,,
\end{aligned}
\end{equation} 
where Graph-Pooling is the max-pooling operation over the RCGA.
At the meantime, we also replace the original token representation $\bm{h}_i$ used in Eq.(\ref{pos-guided ptr}) with the updated one $\bm{h}^{'}_i$.

Finally, we reinforce the first-order biaffine role type detector in Eq.(\ref{biaffine}) with a high-order triaffine attention:
\begin{equation}
\begin{aligned}
\setlength\abovedisplayskip{2pt}
\setlength\belowdisplayskip{2pt}
y^c &=
\text{Sigmoid} ( \left[
\begin{array}{c}
  \bar{\bm{a}}^{o}    \\
    1
\end{array}
\right]^\mathsf{T}
(\bar{\bm{a}}^{r})^\mathsf{T}
\mathbf{W}_{10} \left[
\begin{array}{c}
  \bar{\bm{g}}    \\
    1
\end{array}
\right]  )  \,,
\end{aligned}
\end{equation}
where $\bar{\bm{a}}^{*}$ is the updated term representations.
Intuitively, the role type prediction can be improved in the help of such global-level feature.

\section{Experiment}

\subsection{Setup}

\noindent\textbf{Datasets.}
We experiment with the ORL benchmark, MPQA v2.0.
The dataset has a fixed number of 132 documents for developing.
And for the rest of 350 documents, following the prior work \cite{zhang-etal-2020-syntax,xia-etal-2021-unified}, we take five-fold cross-validation, i.e., splitting 350 documents into 280 and 70 documents as training and testing, respectively.

\noindent\textbf{Implementation.}
We use the pre-trained 300-d GloVe (Pennington et al. \citeyear{pennington-etal-2014-glove}) for word-level embedding, and BERT (base cased version) also is additionally used \cite{devlin-etal-2019-bert}.
The kernel sizes of CNN for character-level representations are [3,4,5].
The hidden size of BiLSTM is 150, and the Stack-LSTM is 300.
Other embedding sizes are all set to 50.
We denote the vanilla neural transition model as `PtrTrans', and the syntax-enhanced version as `SyPtrTrans'.

\begin{table}[!t]
\begin{center}
\resizebox{0.96\columnwidth}{!}{
  \begin{tabular}{lcccc}
\toprule
 & $\mathcal{O}$ &  $\mathcal{O}$-$\mathcal{R}$ & $\mathcal{O}$-$\mathcal{R}$(${hd}$) & $\mathcal{O}$-$\mathcal{R}$($tg$)\\
\midrule 
\multicolumn{5}{l}{$\bullet$ \bf w/o BERT} \\
\quad BiLSTM+CRF$^\dag$ & 52.65 & - & 46.72 & 30.07 \\
\quad  Transition$^\dag$ & 53.04 & -  & 47.02 & 31.45 \\
\quad S{\small PAN}OM$^\dag$  & 58.06 & 43.12 & 52.90 & 32.42  \\
 \quad  PtrTrans & \underline{58.14} &  \underline{43.66} & \underline{53.19}&  \underline{33.23}\\
  \quad  SyPtrTrans & \bf 59.87 & \bf 44.41 & \bf 54.67&  \bf 35.03\\
\hline
\multicolumn{5}{l}{$\bullet$ \bf w/ BERT} \\
\quad S{\small PAN}OM$^\dag$ & 63.71 &  49.89 &58.24& 41.10\\
\quad S{\small PAN}OM+Syn$^\dag$ & - &  \underline{50.46} & \underline{58.46}&  41.82 \\
 \quad  PtrTrans & \underline{63.90} &  50.11 & 58.28&  \underline{41.96} \\ 
  \quad SyPtrTrans & \bf 65.28 &  \bf51.62 & \bf 59.48& \bf 44.04 \\ 
\bottomrule
 \end{tabular}
}
\end{center}
  \caption{
Main results in exact F1 scores.
Baselines with the superscript `$\dag$' are copied from  Xia et al. (\citeyear{xia-etal-2021-unified}).
  }
  \label{main}
\end{table}

\begin{table}[!t]
\begin{center}
\resizebox{1.0\columnwidth}{!}{
  \begin{tabular}{lcccc}
\toprule
\multirow{2}{*}{}&  \multicolumn{2}{c}{Binary F1} & \multicolumn{2}{c}{Proportional F1} \\
\cmidrule(r){2-3}\cmidrule(r){4-5}
& \multicolumn{1}{c}{$\mathcal{O}$-$\mathcal{R}$($hd$)} &\multicolumn{1}{c}{$\mathcal{O}$-$\mathcal{R}$($tg$)}  &\multicolumn{1}{c}{$\mathcal{O}$-$\mathcal{R}$($hd$)}  &\multicolumn{1}{c}{$\mathcal{O}$-$\mathcal{R}$($tg$)}\\
\midrule
\multicolumn{5}{l}{$\bullet$ \bf w/o BERT} \\
\quad BiLSTM+CRF$^\dag$ & 58.22 & \underline{54.98} & - & - \\
\quad  Transition$^\dag$ & \bf 60.93 & \bf 56.44 & - & - \\
\quad S{\small PAN}OM$^\dag$  & 56.47 & 45.09 & 55.62 & 41.65 \\
 \quad  PtrTrans & 57.91 & 46.96& \underline{ 56.88} & \underline{42.82}\\
 \quad  SyPtrTrans & \underline{58.29} & 47.68& \bf 57.12 & \bf 43.63\\
\hline
\multicolumn{5}{l}{$\bullet$ \bf w/ BERT} \\
\quad BERT+CRF$^\dag$ & 55.52 & 50.39 & 46.62 & 34.29 \\
\quad S{\small PAN}OM$^\dag$ & 62.04 & 53.27& 61.20 & 49.88 \\
 \quad  PtrTrans & \underline{62.25} & \underline{53.66} & \underline{61.65} & \underline{50.42} \\ 
  \quad  SyPtrTrans & \bf 63.22 & \bf 55.23&  \bf 62.34 & \bf 52.02 \\ 
\bottomrule
  \end{tabular}
}
\end{center}
  \caption{
  The results in binary and proportional F1 scores.
  }
  \label{binary and proportional}
\end{table}

\noindent\textbf{Baselines.}
We make comparisons with the existing end-to-end ORL baselines. 
1) \textbf{BiLSTM+CRF}: Katiyar et al. (\citeyear{katiyar-cardie-2016-investigating}) propose a joint ORL model based on an enhanced sequence labeling scheme.
2) \textbf{Transition} model for joint ORL (Zhang et al. \citeyear{ZhangWF19}).
3) \textbf{S{\small PAN}OM}: a span-based graph model (Xia et al. \citeyear{xia-etal-2021-unified}).
Also, \textbf{S{\small PAN}OM+Syn} is the version that integrates syntactic constituent features.

\noindent\textbf{Evaluation.}
We follow \citeauthor{xia-etal-2021-unified}(\citeyear{xia-etal-2021-unified}), taking the \textbf{exact F1}, \textbf{binary F1} and \textbf{proportional F1} scores as the evaluating metrics.
We measure the following prediction objects (or subtasks):
1) the opinion terms ($\mathcal{O}$), where a term is correct when both the start and end boundaries of a term are correct;
2) the opinion-role pairs, including the roles of \emph{holder} type ($\mathcal{O}$-$\mathcal{R}$($hd$)), the \emph{target} type ($\mathcal{O}$-$\mathcal{R}$($tg$)), and the overall both two type ($\mathcal{O}$-$\mathcal{R}$).
A pair prediction is correct only when the opinion term, role term and
role type are all correct.
All results of our two models are presented after a significant test with the best baselines, with p$\le$0.05.

\subsection{Results and Discussion}

\noindent\textbf{Main Result.}
Table \ref{main} reports the results of different models.
Firstly let's check the performances with GloVe embedding.
We see that PtrTrans is superior to all baselines, especially outperforming the best-performing baseline, {S{\small PAN}OM}.
Even based on the homologous transition technique, our PtrTrans significantly beats the one of Zhang et al. (\citeyear{ZhangWF19}).
This is largely because our transition system can deal with the term overlap issues; and meanwhile our PointNet design empowers more accurate term extraction.
More crucially, our syntax-enhanced SyPtrTrans model brings the most prominent results against all baselines on total of four subtasks.

Further, with the help of BERT, all systems obtain universally boosted performances.
Also with the integration of external syntax information, the results can be improved.
For example, even the syntax-aware baseline S{\small PAN}OM+Syn surpasses our PtrTrans model on $\mathcal{O}$-$\mathcal{R}$ pair detection.
Nevertheless, our SyPtrTrans system still keeps its great advantage, e.g., giving the best exact F1 scores, i.e., 65.28\% on opinion term extraction and 51.62\% on the $\mathcal{O}$-$\mathcal{R}$.

In Table \ref{binary and proportional} we additionally present the performances under binary F1 and proportional F1 metrics.
We observe that the weak baselines (e.g., Transition, BiLSTM+CRF) with GloVe embedding achieve higher binary F1 scores.
The underlying reasons can be that these models tend to give more extractions on the shorter text fragments, and thus result in higher scores under the `binary measurement'.
In contrast, our systems, due to the equipment of PointNet, can successfully manage the detection of terms, especially including the longer ones.
Detailed analysis is presented later.

\begin{table}[!t]
\begin{center}
\resizebox{0.84\columnwidth}{!}{
  \begin{tabular}{lll}
\toprule
 &  \multicolumn{1}{c}{$\mathcal{O}$-$\mathcal{R}$($hd$) / $\Delta$} & \multicolumn{1}{c}{$\mathcal{O}$-$\mathcal{R}$($tg$) / $\Delta$} \\
\midrule
\multicolumn{1}{l}{SyPtrTrans} &      \bf 59.48 &   \bf 44.04\\
\hdashline
\multicolumn{3}{l}{$\bullet$ \bf Input features} \\
\quad w/o Char ($\bm{x}^{c}_i$)   & 59.18  / -0.30    &   43.75 / -0.29\\
\hdashline
\multicolumn{3}{l}{$\bullet$ \bf Boundary detection features} \\
\quad w/o BDE (Eq.\ref{BDE}) &    59.32 / -0.16 &   43.89 / -0.15\\
\quad w/o POS &   59.20 / -0.28 &   43.48 / -0.56\\
\hdashline
\multicolumn{3}{l}{$\bullet$ \bf High-order Interaction features} \\
\quad w/o Opn & 58.85 / -0.63   & 42.99 / -1.05\\
\quad w/o Dep   &     58.72 / -0.76   & 42.58 / -1.46\\
\quad w/o UDOG  &   58.56 / -0.92   & 42.13 / -1.91\\
\bottomrule
  \end{tabular}
}
\end{center}
  \caption{
  The ablation results (exact F1).
  `w/o Opn' means building UDOG without using the opinion-role structure, and `w/o Dep' without using dependency structure.
  }
  \label{ablation}
\end{table}

\begin{figure}[!t]
\centering
\includegraphics[width=1\columnwidth]{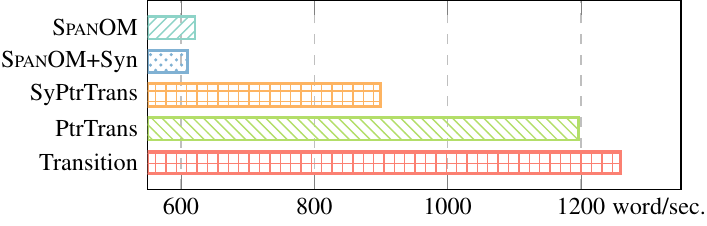}
\caption{
Comparisons on decoding speed.
}
\label{speed}
\end{figure}

\noindent\textbf{Ablation Study.}
We can first learn from Table \ref{main} and Table \ref{binary and proportional} that the BERT contextualized word embedding shows the greatest impacts to the system, demonstrating its extraordinary usefulness to the semantics capturing.
Table \ref{ablation} shows the ablation of our best SyPtrTrans model in terms of other factors.
We can see that the character feature as well as the \emph{Boundary Differential Enhancer} mechanism show some certain impacts to the system, while stopping injecting the POS tag features give worse performances for the term boundary detection.
Looking into the syntax-aided high-order interaction, we observe that the unavailability of either the opinion-role structure or the syntactic dependency structure could lead to sub-optimal task prediction, while the whole removal of UDOG substantially deteriorates the overall performances due to no modeling the interactions among the opinion and role terms.

\begin{figure}[!t]
\centering
\includegraphics[width=0.98\columnwidth]{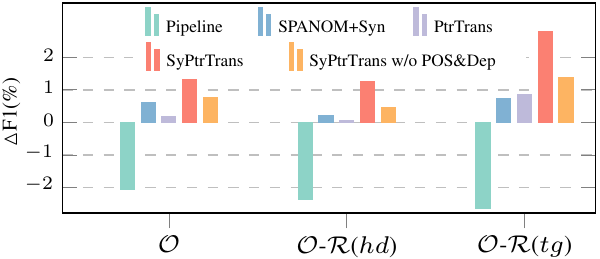}
\caption{
Performance drops of an ORL system comparing to S{\small PAN}OM.
We re-implement a BERT+sequence-labeling model as the \emph{Pipeline} baseline: i.e., first extracting terms and then role type relation classification.
}
\label{interaction}
\end{figure}

\noindent\textbf{Decoding Efficiency.}
In Fig. \ref{speed} we compare the decoding speed of different systems.
We observe that these transition-based methods show very significant model efficiency, due to their linear-time complexity characteristics.
Our PtrTrans achieves 2$\times$ faster decoding than the span-based graph models (i.e., S{\small PAN}OM or with syntax enhancement), due to their iterative enumerating nature.
We can also notice that the Transition baseline (Zhang et al. \citeyear{ZhangWF19}) runs slightly faster than ours, partially due to the more sophisticated design of our model.
Besides, integrating the external syntax information also could considerably degrade the inference speed, i.e., PtrTrans vs. SyPtrTrans.

\noindent\textbf{Study of the Interaction.}
We investigate the impact of interaction mechanisms for ORL, as shown in Fig. \ref{interaction}.
First of all, the pipeline system without any interaction between two stages of subtasks results in a severe performance drop compared to the joint method of S{\small PAN}OM.
Given that the unified ORL modeling naturally entails certain implicit interactions, our PtrTrans still achieves slightly higher results than S{\small PAN}OM or even with syntax aided.
Notedly, we see that with the explicit interaction from co-modeling of opinion-role structure and syntax structure, our SyPtrTrans can significantly outperform those systems with `implicit interaction'.
Even the explicit interaction is merely from the partial opinion structure, i.e., without integrating external knowledge (POS\&dependency), SyPtrTrans still gives higher performances than others.

\begin{figure}[!t]
\centering
\includegraphics[width=0.98\columnwidth]{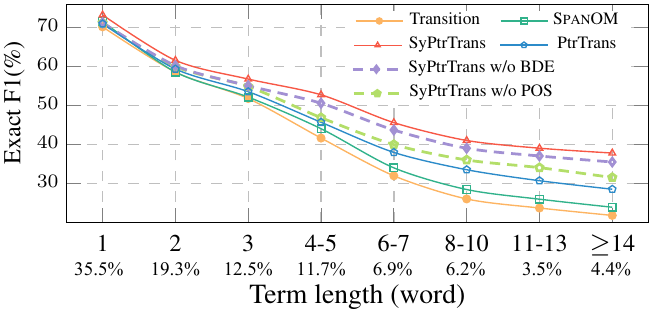}
\caption{
Results of term extraction by varying term lengths.
}
\label{span-length}
\end{figure}

\noindent\textbf{Impact of Term Length.}
We further study the impact of different role term lengths.
From the trends in Fig. \ref{span-length} we see that overall, the performances decrease with the increased term length.
Also, due to span term enumerating characteristics, the span-based graph model S{\small PAN}OM performs slightly better than the Transition baseline.
In the contrast, our transition frameworks with the design of pointer network are more extraordinary than any baseline on handling terms at any length, especially on the longer terms.
More prominently, the additional boundary features from POS contribute a lot to fighting against the challenge of long-length terms, among which the proposed \emph{boundary differential enhancing} mechanism also benefits to some certain extent.

\begin{figure}[!t]
\centering
\includegraphics[width=0.96\columnwidth]{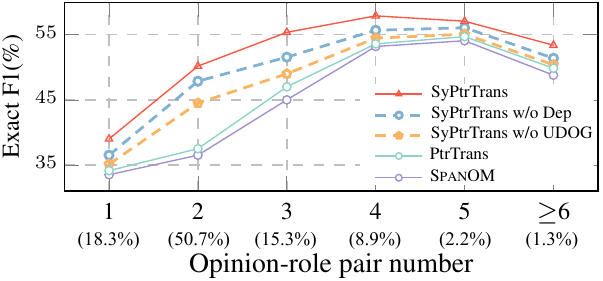}
\caption{
Performances ($\mathcal{O}$-$\mathcal{R}$) with different opinion-role pair number in a sentence.
}
\label{pair-number}
\end{figure}

\vspace{3pt}
\noindent\textbf{Influences of the Opinion-role Pair Numbers.}
We finally in Fig. \ref{pair-number} plot the results with different numbers of opinion-role pairs in a sentence.
We observe the essential pattern of ORL task that, fewer co-existence of opinion and role pairs (i.e., $\le$2) can actually lead to lower recognizing performances of all models, comparing to the results with more opinion-role pairs in sentences.
The main cause is that too few opinion-role pairs result in positive label sparsity, while more pair co-occurrence (i.e., $\ge$4) provides rich training signals and enables better learning.
Under such a circumstance, the model considering rich interactions of opinion-role pairs (e.g., our SyPtrTrans via UDOG) could consequently enhance the task performances, comparing to those without explicit interactions (e.g., PtrTrans and S{\small PAN}OM).
Also from the trends, we see that the co-modeling of the syntactic dependency structure for building the UDOG is quite important to the high-order interaction.

\section{Conclusion}

In this work we present a novel unified opinion role labeling framework by implementing a neural transition architecture with a pointer network (PointNet).
The system enables more accurate predictions on longer terms, and solving the overlap issue meanwhile advancing in a linear-time decoding efficiency.
We then enhance the PointNet for boundary detection by integrating linguistic POS tag features.
We further explore an explicit interaction between the opinion and role terms by co-modeling the syntactic dependency structure and the partial opinion-role structure as a unified graph, which is further encoded via a novel relation-centered graph aggregator.
Experimental results demonstrate that our vanilla transition system outperforms the top-performing baseline, meanwhile achieving 2$\times$ faster decoding.
By capturing rich high-order features via explicit interaction, our syntax-enhanced system obtains new state-of-the-art results on the benchmark dataset.

\newpage

\section*{Acknowledgments}

This work is supported by the National Natural Science Foundation of China (No.61772378, No. 62176187), 
the National Key Research and Development Program of China (No. 2017YFC1200500), 
the Research Foundation of Ministry of Education of China (No.18JZD015).

\bibliography{aaai22}
\end{document}